\documentclass[Afour, sageh, times]{sagej}
\usepackage{moreverb, url}
\usepackage[
    colorlinks,
    bookmarksopen,
    bookmarksnumbered,
    citecolor=red,
    urlcolor=red
]{hyperref}
\usepackage[acronym]{glossaries}
\usepackage{siunitx}
\usepackage{soul}
\usepackage{amsmath}
\usepackage{nicematrix}
\usepackage{dirtree}
\usepackage{orcidlink}

\newcommand\BibTeX{{\rmfamily B\kern-.05em \textsc{i\kern-.025em b}\kern-.08em
T\kern-.1667em\lower.7ex\hbox{E}\kern-.125emX}}
\def\volumeyear{2024}

\newcommand{\etal}{\emph{et~al.}}
\newcommand{\secref}[1]{Section~\ref{#1}}
\newcommand{\figref}[1]{Figure~\ref{#1}}

\begin{document}
\runninghead{Mansour \etal}
\title{eCARLA-scenes: A synthetically generated dataset for event-based optical flow prediction}
\author{Jad Mansour\affilnum{1}~\orcidlink{0009-0007-8559-6753}, Hayat Rajani\affilnum{1}~\orcidlink{0000-0002-2541-2787}, Rafael Garcia\affilnum{1}~\orcidlink{0000-0002-1681-6229}, and Nuno Gracias\affilnum{1}~\orcidlink{0000-0002-4675-9595}}
\affiliation{\affilnum{1}Computer Vision and Robotics Research Institute (ViCOROB), \\ University of Girona (UdG), Spain}
\corrauth{Jad Mansour, Edifici CIRS, Parc Científic i Tecnològic UdG, \\ Carrer Pic de Peguera 13, 17003 Girona, Spain}
\email{jad.mansour@udg.edu}

\newacronym{ebc}{EBC}{Event-based Camera}
\newacronym{uav}{UAV}{Unmanned Aerial Vehicle}
\newacronym{auv}{AUV}{Autonomous Underwater Vehicle}
\newacronym{rov}{ROV}{Remotely Operated Vehicle}
\newacronym{ann}{ANN}{Artificial Neural Network}
\newacronym{cnn}{CNN}{Convolutional Neural Network}
\newacronym{snn}{SNN}{Spiking Neural Network}
\newacronym{if}{IF}{Integrate-and-Fire}
\newacronym{lif}{LIF}{Leaky Integrate-and-Fire}
\newacronym{aee}{AEE}{Average Endpoint Error}
\newacronym{aae}{AAE}{Average Angular Error}

\begin{abstract}
    The joint use of event-based vision and \glspl*{snn} is expected to have a large impact in robotics in the near future, in tasks such as, visual odometry and obstacle avoidance. While researchers have used real-world event datasets for optical flow prediction (mostly captured with \glspl*{uav}), these datasets are limited in diversity, scalability, and are challenging to collect. Thus, synthetic datasets offer a scalable alternative by bridging the gap between reality and simulation. In this work, we address the lack of datasets by introducing \emph{eWiz}, a comprehensive library for processing event-based data. It includes tools for data loading, augmentation, visualization, encoding, and generation of training data, along with loss functions and performance metrics. We further present a synthetic event-based datasets and data generation pipelines for optical flow prediction tasks. Built on top of \emph{eWiz}, \emph{eCARLA-scenes} makes use of the \emph{CARLA} simulator to simulate self-driving car scenarios. The ultimate goal of this dataset is the depiction of diverse environments while laying a foundation for advancing event-based camera applications in autonomous field vehicle navigation, paving the way for using \glspl*{snn} on neuromorphic hardware such as the Intel Loihi.
\end{abstract}

\maketitle

\section{Introduction}
\label{sec:introduction}
    \Glspl*{ebc}, also known as dynamic vision sensors or neuromorphic cameras, are imaging sensors that operate differently from traditional frame-based cameras. Inspired by the exceptional motion perception abilities of winged insects, these cameras respond asynchronously to brightness or intensity changes with microsecond resolution \citep{camera1Lichtsteiner08, camera2Brandli14}. \Glspl*{ebc} offer significant advantages over traditional frame-based cameras: no motion blur, high dynamic range, high temporal resolution, and low latency \citep{camera3Gallego20}. These unique properties make the \gls*{ebc} an ideal sensor to analyze dynamic scenes characterized by fast motion and rapid changes in lighting conditions.

    \Glspl*{ebc} and optical flow prediction are two complementary concepts that play an integral part in motion analysis in computer vision. Traditionally, optical flow prediction has relied on analyzing a sequence of frames to infer motion patterns \citep{flowShah21}. However, \glspl*{ebc} present a new paradigm by providing an asynchronous and sparse data representation of data in the form of a stream of events \citep{camera3Gallego20}. As such, many methodologies have been developed for event-based processing, though primarily for relatively simple scenes containing well-defined shapes, edges, and corners \citep{icpKueng16, probabilityZhu17, ekltGehrig20}. However, these methods have proven to be inefficient and computationally expensive in complex scenes.

    An alternative approach is the use of \glspl*{ann} for processing event-based data, which has demonstrated remarkable success in modeling and understanding the nature of these biological vision systems. One significant drawback of \glspl*{ebc}, however, is their incompatibility with conventional \gls*{ann} models, such as \glspl*{cnn}. This necessitates the use of intricate data encoding schemes to convert the sparse asynchronous stream of events into a frame-based representation that can be handled by \glspl*{cnn}, as demonstrated by researchers \citep{evflownetZhu18, egoflownetZihao19, biflowWan22} who have implemented U-net \citep{unetWeng15} like architectures for event-based optical flow estimation. However, such compression of event streams into frame-based representations inevitably results in loss of information, thereby limiting the potential of \glspl*{ebc}. As a result, they demand newer paradigms specifically optimized for handling such data. A \gls*{snn} is one such brain-inspired computing model that relies on so-called spiking neurons to process information \citep{snnTavanaei19}. Spiking neurons are inherently designed to process temporal patterns, making them naturally compatible with the asynchronous stream of events. An interconnection of such spiking neurons in an \gls*{snn} allows them to effectively extract spatial features, enabling a comprehensive framework for encoding dynamic visual motion over time.

    Nonetheless, there is still a lack of sufficiently diverse and well-labeled datasets to facilitate model training. Generating real-world data is both expensive and time-consuming, often requiring specialized equipment to obtain corresponding ground truth data\textemdash an alternative to the tedious process of manual annotations by human experts. For instance, datasets such as MVSEC \citep{mvsecZhu18} and DSEC \citep{dsecGehrig21}, utilize 3D LiDARs to create a 3D map of the environment using a LiDAR odometry and mapping algorithm. This reconstructed environment is then project onto the \gls*{ebc} sensor to generate a dense optical flow for use as ground truth during training. However, the resulting optical flow includes some inaccuracies due to clock synchronization issues and often tends to miss information, failing to fully represent the event sequences captured by the sensor. As such, these real-world datasets may not be ideal for model training or evaluation, which stresses the need for a synthetic alternative. In fact, due to the nature of \glspl*{ebc}, synthetic event-based data may translate well to real-world data. The features extracted from synthetic frame-based camera datasets often exhibit stark disparities from the features extracted from real-world data. These disparities typically arise from differences in lighting conditions, camera noise, and environmental complexity in real-world settings which are challenging to replicate accurately in synthetic data. Such variations can lead to shifts in pixel intensities, color distribution and texture details, resulting in significant differences between the features extracted from synthetic and real-world images. In contrast, \glspl*{ebc} only trigger events when detecting brightness changes, producing an events image that only has 2 values, a positive or a negative polarity. As such, unlike frame-based images that rely on a static grid of pixels with values typically ranging between 0 and 255, the output of \glspl*{ebc} is simpler, leaving more room for error when transitioning from simulation to the real-world. Hence, we present the \emph{eCARLA-scenes} dataset, a synthetic event-based optical flow dataset generated by simulating diverse environments under different weather conditions with the CARLA simulator \citep{carlaDosovitskiy17}. We also provide an easy-to-use pipeline for creating custom datasets with pre-synchronized optical flow displacements and grayscale images. The dataset is hosted on Zenodo (DOI: \url{10.5281/zenodo.14412251}) and the data generation pipeline is publicly available under the following link \url{https://github.com/CIRS-Girona/ecarla-scenes}.

    Finally, we also introduce \emph{eWiz}, a library that is composed of all the tools necessary for processing event-based data, ranging from data loading and manipulation, to augmentation and visualization, to encoding and training data generation, together with various loss functions and performance metrics. We plan to add more functionalities and incorporate more areas of applications as our work on event-based processing progresses. The library is openly available under the following link \url{https://github.com/CIRS-Girona/ewiz}.

    In summary, our main contributions are:
    \begin{itemize}
        \item The generation of \emph{eCARLA-scenes}, a diverse event-based optical flow dataset using the CARLA simulator, along with the \emph{eCARLA} pipeline, which simplifies the process of extending the datasets.
        \item The creation of \emph{eWiz}, an all inclusive framework for event-based data processing and manipulation.
    \end{itemize}

    The remainder of this paper is organized as follows. \secref{sec:related_work} presents an overview of other publicly available datasets for event-based optical flow prediction. \secref{sec:data_simulation} and \secref{sec:data_generation} describe the details of the simulation setup and sensor configurations used for our datasets, along with the data generation pipelines. \secref{sec:data_processing}, on the other hand, presents details on the \emph{eWiz} library for additional processing of the datasets for visualization, analysis and model training. \secref{sec:data_repository} gives an overview of how the datasets are structured. Finally, \secref{sec:conclusion} concludes this study and presents directions for future work.

\section{Related Work}
\label{sec:related_work}
    In the literature, only a limited number of real-world event-based datasets have been employed for developing and evaluating event-vision algorithms and \glspl*{ann} targeting optical flow prediction. Among these, the widely used MVSEC dataset \citep{mvsecZhu18} stands out, offering two distinct scenarios: a stereo pair of \glspl*{ebc} mounted on an \gls*{uav} in an indoor environment, and the same setup mounted on a car in an outdoor setting. The ground truth optical flow for indoor scenes is derived using a motion capture system, while for outdoor scenes, it is obtained through a combination of GPS data and scene reconstruction techniques. Building upon MVSEC, the DSEC dataset, introduced by \citet{dsecGehrig21}, presents significant improvements. It features driving scenarios with displacement field magnitudes reaching up to 210 pixels and a camera resolution three times higher than that of MVSEC. While these datasets have been instrumental in advancing event-based vision research, they are not without limitations. The generation of real-world data is both resource-intensive and time-consuming, often necessitating specialized equipment to capture accurate ground truth information. This reliance on costly setups further underscores the challenges of producing high-quality datasets in this domain. Moreover, the MVSEC dataset exhibits notable limitations in the accuracy of its ground truth optical flow. The ground truth was generated using a Velodyne LiDAR, which relies on a 3D reconstruction of the environment to estimate optical flow. The dataset suffers from additional challenges due to the mechanical setup. Specifically, the vibrations of the car during forward motion induce a wobbling effect in the event-based sensor. This wobble results in the sensor capturing significant up-and-down motion, whereas the ground truth optical flow predominantly reflects forward motion. This mismatch between the motion perceived by the event-based sensor and the ground truth optical flow can introduce discrepancies in training and evaluating models, especially for applications that rely on precise motion estimation.

    These challenges contribute to the scarcity of real-world event-based datasets, prompting researchers to increasingly turn to synthetically generated datasets. Leveraging advanced simulators, synthetic datasets provide a scalable and cost-effective alternative, enabling the generation of diverse and accurately labeled data. ESIM for example, \citet{esimRebecq18}, being one of the most accurate in the literature, makes use of an adaptive rendering scheme that only samples frames when necessary. BlinkSim, proposed by \citet{blinksimLi23}, is another \gls{ebc} simulator that includes a powerful rendering engine capable of building thousands of scenes with different objects and motion patterns at high frequency. This simulator gave rise to the BlinkFlow dataset, which improves generalization performance of state-of-the-art methods by 40\% on average. Another synthetic dataset is the MDR created by \citet{mdrLuo23}. It makes use of Blender to create indoor and outdoor 3D scenes and includes diverse camera motions and high frame-rate videos between images.

    Although synthetic optical flow datasets for event-based vision are already available, most are not specifically designed to reflect the motion dynamics or characteristics of any particular vehicle type. In contrast with MDR, ESIM, and BlinkFlow, which randomly move the sensor around, our work focuses on applications tailored to autonomous field vehicles. We include example of vehicles driving forward, backward, turning, and swaying. We specifically use \emph{CARLA} simulator \citep{carlaDosovitskiy17}, a renowned driving simulator which includes an \gls*{ebc} sensor. Build on top of Unreal Engine 4, CARLA provides accurate rendering in terms of lighting and reflection when compared with other simulators. ESIM uses an OpenGL rendering engine \citep{openglOpengl14}, making it much more complex to generate realistic scenes, especially for car driving scenarios. Moreover, the use of Blender \cite{blenderBlender23} for MDR requires intense knowledge of 3D scene creation to correctly simulate real-world lighting conditions. With CARLA on the other hand, we provide an easy to use \emph{scenario creator} with which driving sequences can be easily recorded. These scenarios can then be easily played back with our \emph{scenario reader} pipeline in which we can set the sensor location and refresh rate, whether for grayscale, events, or optical flow. As such, the data is not only reproducible but also extensible. Finally, we provide the \emph{eCARLA-scenes} dataset taken with our pipeline containing diverse scenes (urban and rural areas), vehicle movements, and weather conditions (sunny, foggy, sunset). Additionally, our dataset includes both static and dynamic scenarios to align with state-of-the-art event-based optical flow loss functions, such as contrast maximization \citep{mcGallego2018, mcGallego2019}. The static scenarios are devoid of moving objects, ensuring compatibility with these loss functions, while the dynamic scenarios feature a variety of configurations, including moving traffic, pedestrians, or a combination of both.

    Finally, although a few open-source tools are available for specific event-based data processing tasks, such as the work by \cite{mcShiba2022}, no comprehensive library currently integrates these capabilities into a unified framework. To address this gap and streamline future development, we present \emph{eWiz}\textemdash an all-in-one solution for event-based data manipulation.

\section{Simulation Setup}
\label{sec:data_simulation}
    This section describes the simulation environments, sensor and vehicle configurations, and the data collection approach for generating the event-based optical flow prediction dataset: \emph{eCARLA-scenes}, tailored for autonomous field vehicles. Both the datasets consist of event streams, grayscale images, and the corresponding ground truth optical flow. Additionally, they utilize an optimized format provided within the \emph{eWiz} framework to facilitate efficient data storage, access, and processing. Further details on this optimized format are provided in \secref{sec:data_processing}.

    \subsection{CARLA Dataset}
        \emph{eCARLA-scenes} was created using the CARLA simulator \citep{carlaDosovitskiy17}, with a deliberate selection of versatile and progressively challenging environments to simulate various features and conditions that an \gls*{ebc} sensor might encounter in real-world scenarios. Specifically, we selected the following four preset environments from CARLA:
        \begin{itemize}
            \item \emph{Town10HD}, a downtown urban environment with skyscrapers, residential buildings, and an ocean promenade which present simple features and well-defined edges.
            \item \emph{Town07}, a rural scene with flora and foliage that showcase examples of natural elements with complex features and edges.
            \item \emph{Town04}, a rural environment characterized by a wide infinity-shaped highway.
            \item \emph{Town02}, a small simple town with a mixture of residential and commercial buildings.
        \end{itemize}
        We also take into account different weather conditions to make the scenes more realistic using CARLA's weather presets. The \emph{ClearNoon} and \emph{ClearSunset} presets simulate clear weather conditions allowing all objects in the scene to be seen distinctly. The primary difference between them is the position of the sun in the sky: at sunset, with the sun positioned in front of the sensor, extreme brightness values are generated, saturating both ends of the brightness spectrum. Additionally, atmospheric scattering in such conditions decreases brightness disparities across frames. We also used the \emph{CloudyNoon} and \emph{CloudySunset} presets, which create a more dimly lit environment where certain edges can appear to be less defined than in sunny conditions. Finally, the \emph{CloudyNoon} weather preset, specifically for \emph{Town04}, introduces light scattering due to fog, which has a notable impact on the amount of events captured by an \gls*{ebc}. It reduces the intensity differences between successive frames, thereby decreasing the number of events triggered at a given threshold.

\begin{figure}[!ht]
            \includegraphics[width=0.45\textwidth]{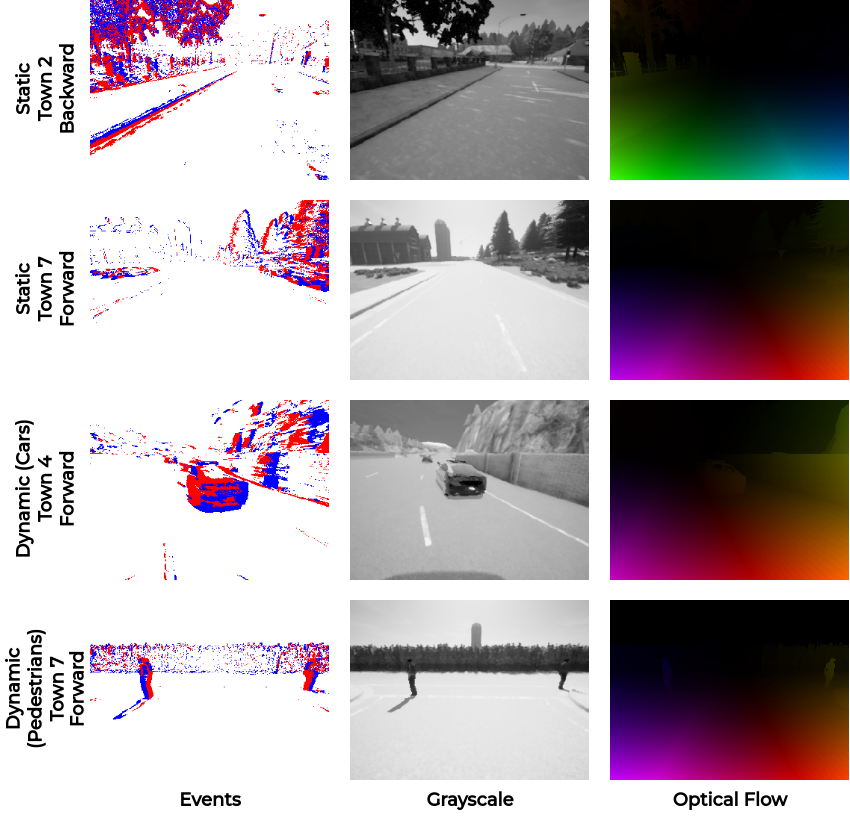}
            \caption{Captured data examples for \emph{eCARLA-scenes}, showcasing events, grayscale images, and optical flow data}
            \label{fig:carla_dataset}
        \end{figure}

        We begin with a static scene\textemdash a simple, empty environment with no dynamic objects such as pedestrians or other vehicles\textemdash and progressively increase the dataset's complexity by adding dynamic elements, starting with a few pedestrians and then including other vehicles. As a result, the dataset contains both static and dynamic sequences with varying scene complexity. We collected a total of \textbf{31} sequences. \figref{fig:carla_dataset} depicts some examples of the generated scenes.

        For data collection, a vehicle equipped with an event-based camera, grayscale camera, and optical flow sensor was used. All sensors were mounted at the same front-view location, and have the same resolution of $260\times346$ pixels. The simulation and the \gls*{ebc} sensor were run at \SI{1000}{\hertz}, while the grayscale camera and the optical flow sensor ran at \SI{25}{\hertz}. Event thresholds were set at 0.5 for positive events and 0.4 for negative events. Each sequence was about \SI{59}{\second} in length, during which the vehicle was manually controlled with arrow keys. Each sequence captures a unique driving scenario, with the vehicle randomly moving forward or backward, or swaying. The throttle speed was also randomly varied to produce sequences that included both slow and full-throttle movement. The optical flow sensor, on the other hand provided accurate dense optical flow ground truth, as opposed to real-world datasets. To avoid network bias, sequences were recorded with bi-directional flow, where the front-facing sensors would record data as the vehicle moved in both forward and backward directions, in contrast to the datasets mentioned in the literature.

        It is important to note that CARLA’s optical flow sensor returns optical flow as a displacement matrix from frame $t+1$ to frame $t$. Consequently, the flow direction needs to be inverted to align with standard metrics and losses, which are designed to support forward optical flow. Therefore, we invert CARLA’s optical flow using the method proposed by \citet{inverseflowSanchez13} during dataset generation.

\section{Data Generation Pipelines}
\label{sec:data_generation}
    \emph{eCARLA-scenes} can be extended upon by using their respective data generation pipelines. These pipelines are based on the \emph{eWiz} library. Consequently, all data generated using these pipelines is saved in the \emph{eWiz} format.

    The \emph{eCARLA-scenes} pipeline is designed to facilitate user-friendly and efficient synthetic data generation through a modular structure. It consists of several interconnected components, each responsible for a specific aspect of the simulation process. The main modules include the \emph{spawn} module, \emph{control} module, \emph{sensor} module, \emph{game} module, \emph{extract} module, and \emph{sync} module. The \emph{spawn} module initializes the environment by spawning the main vehicle, along with any dynamic elements such as pedestrians or other vehicles. Users can customize the environment by selecting the vehicle type and adjusting the number of pedestrians and vehicles to create dynamic driving scenarios. The \emph{control} module allows the user to interact with the simulation through the keyboard, enabling control of the main vehicle within the selected map for any desired duration. The \emph{sensor} module is responsible for defining and configuring the sensors in the simulation. Users can specify the sensor type, name, resolution, refresh rate, and other parameters. The module also allows users to adjust the sensor's pose relative to the main active vehicle via the transform option. The \emph{game} module handles image rendering, while the \emph{extract} module is responsible for extracting and saving data using \emph{eWiz} data writers, storing all recorded information in the \emph{eWiz} format. The \emph{sync} module ensures that the sensors' data is synchronized based on their individual refresh rates. This synchronization guarantees that the data collected during the simulation is temporally consistent. Additionally, the pipeline includes two main scripts: the \emph{scenario creator} and \emph{scenario reader}. The scenario creator allows users to design driving scenarios by selecting their desired map and generating the main vehicle's path. The scenario reader is tasked with acquiring sensor data, where users can define the desired sensors. The reader also ensures proper synchronization of sensors and allows users to replay recorded scenarios, adjusting the simulation's refresh rate and weather conditions as needed.

\section{Data Processing}
\label{sec:data_processing}
    The \emph{eCARLA-scenes} dataset was generated with \gls*{ann} training and inference in mind. To streamline this process, we created \emph{eWiz}, a Python-based library that can be installed with PyPi \citep{pipyInstaller}, serving as an all-inclusive framework for event-based data manipulation, processing, visualization, and evaluation. \emph{eWiz} is designed for seamless integration with libraries such as PyTorch \citep{pytorchLibrary} and Tonic \citep{tonicLibrary}, making it suitable for ML-oriented pipelines. Additionally, it offers compatibility with existing datasets and supports both event-based and frame-based data. The following subsections present an overview of the various functionalities currently offered by \emph{eWiz}. For further details, the readers are encouraged to refer to the official documentation.
    \begin{figure}[!ht]
        \includegraphics[width=0.45\textwidth]{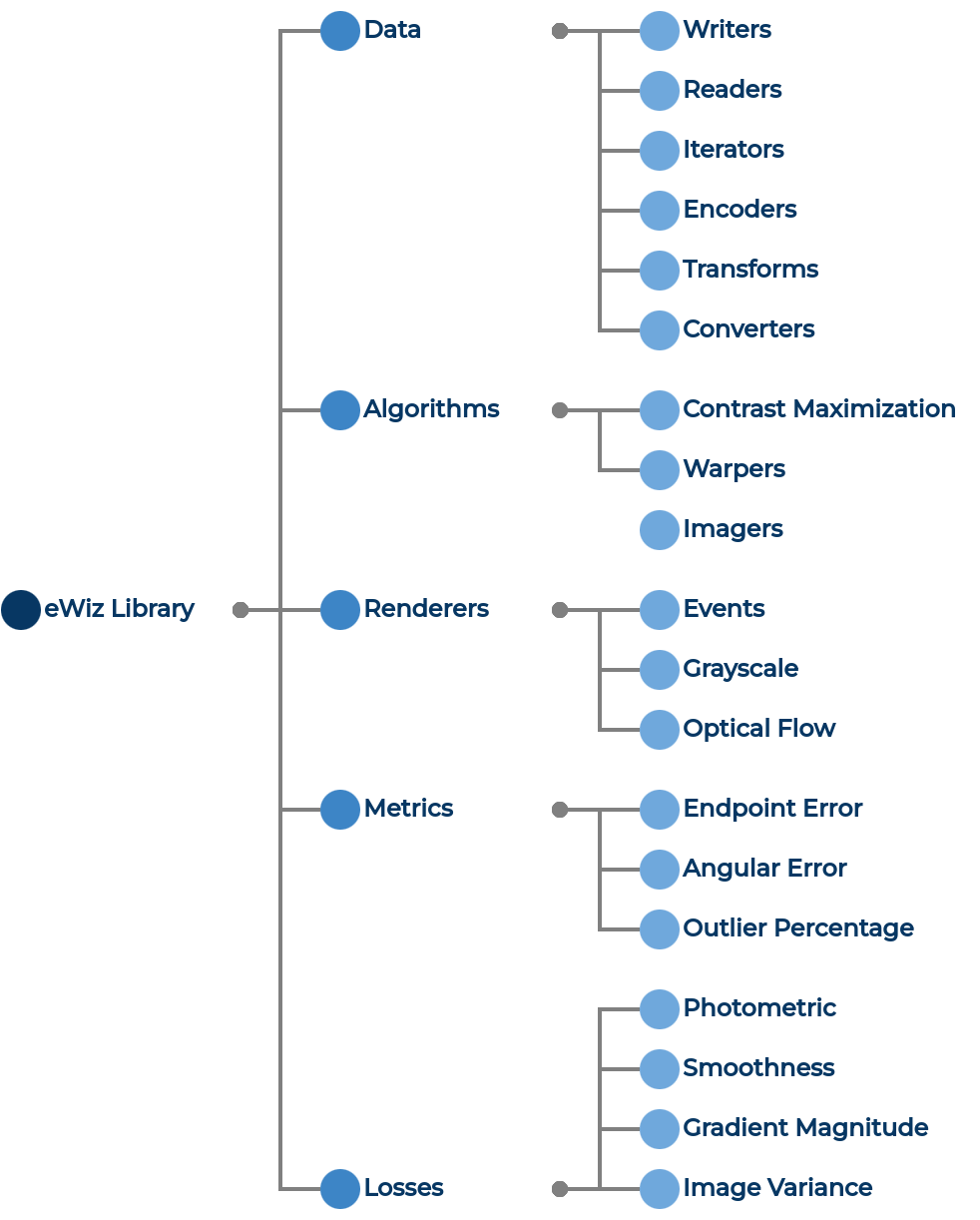}
        \caption{Currently implemented modules for the \emph{eWiz} library}
        \label{fig:ewiz_modules}
    \end{figure}

    \subsection{Data Access and Storage}
        Due to the high temporal resolution and data throughput of \glspl*{ebc}, event-based data requires substantial storage space. Additionally, slicing data between any two timestamps requires time-consuming search algorithms due to the large volume of events in the sequence. Libraries like NumPy \citealp{numpyNumpy} also face high memory demands, as they typically require loading entire sequences into memory before slicing or processing. The \emph{eWiz} format addresses these challenges with an optimized approach to data management and manipulation. Firstly, we use Blosc \citep{bloscCompression}, a high-performance compressor optimized for binary data, effective for compressing numerical arrays. To further reduce data size, we constrain the possible values for every event stream. For instance, pixel coordinates are stored as 16-bit unsigned integers, while polarities are stored as boolean values. Next, we use the \emph{h5py} library \citep{h5pyLibrary} read and write data as HDF5 files \citep{hdf5Format}, which supports large and complex data structures, and allows slicing without the need to completely load the data into memory. Finally, to reduce loading times, we pre-compute the mappings between timestamps and data indices during the data writing process. This includes binary search to map timestamps (in $ms$) to event, grayscale, and flow indices, as well as direct mappings of event indices to their grayscale and flow counterparts. Additionally, optical flow data is automatically synchronized between specified timestamps by interpolating and accumulating flow displacements as needed.

        \emph{eWiz} provides an easy to use readers and writers to facilitate this process. The data reader uses the precomputed time mappings to read the desired indices of events based on the start and end timestamps and loads them to memory. It also loads the corresponding grayscale images that were captured at the beginning and at the end of the event stream along with the synchronized optical flow. The user has the possibility to stride over the events, by either using event indices, timestamps, or grayscale images indices. The data reader goes hand in hand with the included data iterators which allow to sequentially stride over the data. Such functionality allows the user to obtain event-based data sequentially, without the need for manual indexing.

        \emph{eWiz} also aims to support a multitude of \glspl*{ebc} and ensure compatibility with publicly available datasets. We include several data converters for transforming these data types into the \emph{eWiz} format. Currently, we support the DAVIS346 \citep{davis346} and PROPHESEE \citep{prophesee} cameras, along with popular datasets such as MVSEC \citep{mvsecZhu18} and DSEC \citep{dsecGehrig21}.
    
    \subsection{Data Encoding and Augmentation}
        Traditionally, \gls*{ann}-based approaches \citep{evflownetZhu18, egoflownetZihao19, biflowWan22} require the \gls*{ebc} output to be converted from raw streams of events data to 2-channel event images, where one channel represents positive polarities and the other the negative polarities.  To support this, \emph{eWiz} provides different encoders. Currently, two encoding schemes are implemented: Gaussian \citep{gaussianDing22} and event counting \cite{countNguyen19}. The former computes pixel values by summing each event’s polarity weighted by a normalized Gaussian kernel based on the event timestamp and a scaling factor $\lambda$ for each bin. Whereas, the latter simply groups events into fixed time intervals, with pixel values computed by summing the polarities of all events within each bin.
    
        \emph{eWiz} also provides numerous event-based temporal and spatial data augmentations, including techniques such as time warping, noise injection, event flipping, and spatial transformations, all of which are fully compatible with PyTorch and Tonic. These augmentations enable enhanced model generalization by simulating diverse real-world conditions, making them ideal for training robust event-based neural networks.

    \subsection{Data Visualization}
        \emph{eWiz} includes several rendering modules for visualizing the acquired data. The primary renderer replays captured sequences, overlaying event images on top of grayscale images for an intuitive view of the data. Additionally, the library offers separate visualizers for event-based data, grayscale images, and optical flow, with the capability to save rendered sequences. Users can visualize the data as: video sequences, individual (encoded) frames, or 3D volumes of event streams with interleaved frames.

    \subsection{Algorithms}
        \emph{eWiz} not only supports data processing for \gls*{ann}-based approaches but also algorithms, such as contrast maximization \citep{mcShiba2022}, that directly operate on raw event streams. Contrast maximization involves warping the event stream based on a proposed optical flow, iteratively optimizing the optical flow field to maximize the contrast computed between the warped and the original event streams. Higher contrast, indicated by well-defined edges, signals improved alignment of events. As such, \emph{eWiz} also provides warping tools that allow the implementation of various motion models based on the camera movement type. In the long run, \emph{eWiz} will support not only optimization-based algorithms but also deep learning-based and \gls*{snn}-based alternatives. We plan to add more functionalities, such as, ground truth generation via optimization, and \gls*{snn}-based training algorithms.

    \subsection{Loss Functions and Evaluation Metrics}
        \emph{eWiz} supports multiple loss functions for training \glspl*{ann}, currently focused on optical flow prediction. For instance, the photometric loss \citep{evflownetZhu18}, as a self-supervised approach to minimize pixel-wise intensity differences between two grayscale images encompassing an event stream, using the predicted flow field to warp the event stream and optimize alignment. This loss can be combined with Charbonnier loss, a smooth approximation of the L1 loss that is more sensitive to large outliers, as well as a smoothness loss serving as a regularizer to address the aperture problem and ensure continuous flow across sparse event-based data. In addition, the library includes losses specifically designed for the contrast maximization algorithm \citep{mcGallego2018, mcGallego2019} to evaluate warped event streams. These include image variance and gradient magnitude loss functions, along with their improved variants presented by \citet{mcShiba2022}, where the losses are normalized and applied in a multi-focal manner. To further enhance performance, \emph{eWiz} also provides a regularizer that smooths the predicted flow across the image, helping address the sparsity of event-based data and ensuring continuous flow, ultimately improving convergence during training.

        Moreover, \emph{eWiz} can be used for evaluating optical flow prediction performance metrics through the \gls*{aee} \cite{evflownetZhu18}. The \gls*{aee} is defined as the distance between the endpoints of the predicted $(u_{pred}, v_{pred})$ and ground truth $(u_{gt}, v_{gt})$ flow vectors, it is averaged over the number of pixels. We also use the \gls*{aae}, defined as the angular difference between the prediction and ground truth flows. Due to the sparsity of event-based data, we only compute these metrics for pixels in which events occurred. Another used metric is the outlier percentage. This metric refers to the percentage of pixels in the image with an endpoint error higher than a certain value. We define \emph{XPE} the overall outlier percentage of pixels with endpoint errors greater than \emph{X}. The \emph{metrics} module of \emph{eWiz} encapsulates all these metrics and keeps tracks of them for algorithmic or \gls*{ann} evaluation. As such, the library can return the overall errors for an entire dataset.

\section{Dataset Structure}
\label{sec:data_repository}
\begin{figure}[!ht]
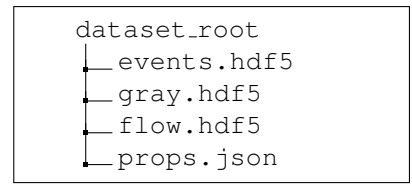

        \centering
        \framebox[0.3\textwidth]{%
            \begin{minipage}{0.25\textwidth}
                \dirtree{%
                    .1 dataset\_root.
                    .2 events.hdf5.
                    .2 gray.hdf5.
                    .2 flow.hdf5.
                    .2 props.json.
                }
            \end{minipage}
        }
        \caption[data structure]{\centering Structure of the data repository.}
        \label{fig:data_structure}
    \end{figure}

    \emph{eCARLA-scenes} follow the \emph{eWiz} format, which stores data in HDF5 files \citep{hdf5Format}. Specifically, events.hdf5 holds the event streams, gray.hdf5 contains grayscale image sequences associated with each event stream, and flow.hdf5 stores the optical flow information. Metadata and configuration details, including sensor specifications and recording parameters, are saved in props.json. This structure facilitates straightforward access to each data type, improving data handling and compatibility with various processing pipelines. The directory structure is summarized in \figref{fig:data_structure}.

    The official documentation provides detailed information on the contents of each HDF5 file. However, users can easily access or write data through the readers and writers included in \emph{eWiz} without needing to handle these details directly.

\section{Conclusion}
\label{sec:conclusion}
    In conclusion, this work introduces the \emph{eCARLA-scenes} dataset, along with a user-friendly pipeline to extend and adapt the dataset. These synthetic event-based optical flow datasets were created to make-up for the lack of labeled event-based datasets in the literature. \emph{eCARLA-scenes} is specifically tailored for driving scenarios in which the vehicle sways and moves forward and backward in different weather conditions. We also present \emph{eWiz}, an all-inclusive event-based data manipulation library that includes an optimized event data format. The library includes data converters, writers, readers, and processing modules that prioritize data management, compression, and optimization. It also includes loss functions, metrics, contrast maximization algorithms, and visualizers that make working with these datasets easier. All in all, \emph{eWiz}, and \emph{eCARLA-scenes} are made to streamline research related to \glspl*{ebc} and motion estimation for autonomous vehicles by offering easy-to-use libraries to generate, process, and evaluate event-based datasets.

\begin{funding}
    J.M. was supported by the Joan Oró Grant no. 2024 FI-2 00762. The work was also supported in part by the Spanish government through the SIREC project no. PID2020-116736RB-IOO.
\end{funding}

\begin{dci}
    The authors declared no potential conflicts of interest with respect to the research, authorship, and/or publication of this article.
\end{dci}

\begin{credit}
    \textbf{Jad Mansour:} Conceptualization, Methodology, Software, Data Curation, Writing - original draft; \textbf{Hayat Rajani:} Conceptualization, Methodology, Supervision, Writing - original draft; \textbf{Rafael Garcia:} Methodology, Supervision, Writing - review \& editing; \textbf{Nuno Gracias:} Project Administration, Writing - review \& editing.
\end{credit}

\bibliographystyle{SageH}
\bibliography{references.bib}
\end{document}